%% file: ms.tex
\documentclass[a4page,twocolumn,9pt]{article}  

\usepackage[left=1.2cm,right=1.0cm,top=1.5cm,bottom=1cm]{geometry}

\input mypackages

\pdfoutput=1
\title{Head and eye egocentric gesture recognition \\ for human-robot interaction using eyewear cameras
\thanks{Work supported by project UJI-B2018-44 from \textsl{Pla de promoci\'o de la investigaci\'o de la Universitat Jaume I, Castell\'o, Spain}. The financial support for the research network with code RED2018-102511-T, from \textsl{Ministerio de Ciencia, Innovaci\'on y Universidades}, is acknowledged. }}
\author{Javier Marina-Miranda$^{1}$ and V. Javier Traver$^{2}$}
\date{$^{1}$Universitat Jaume I, E12071-Castell\'on, Spain \\
$^{2}$Institute of New Imaging Technologies, Universitat Jaume I \\
       E12071-Castell\'on, Spain \\
        {\tt\small vtraver@uji.es}}

\begin{document}

\input mycommands
\input statement
\maketitle
\pagestyle{empty}

\begin{abstract}
\input abstract
\end{abstract}

\input introduction

\input methodology
\input experiments
\input discussion
\input conclusions

\section*{ACKNOWLEDGMENT}
We are grateful to the participants in the video capture sessions.


\bibliographystyle{abbrv}
\bibliography{ms}
\balance

\end{document}

%% file: mypackages.tex
\usepackage[utf8]{inputenc} 
\usepackage[hidelinks]{hyperref} 
\usepackage{graphicx} 
\usepackage{xcolor} 

\usepackage{tikz}
\usepackage{ifthen}
\usetikzlibrary{matrix,calc}

\usepackage{subcaption}
\usepackage{multirow}
\usepackage{comment}
\usepackage{sidecap}

\usepackage{balance}

\usepackage{amsmath,amssymb}

%% file: mycommands.tex
\newcommand{\xavi}[1]{\textcolor{purple}{\scriptsize \textsf{#1}}}

\newcommand{\neutral}{\textsf{Neutral}}
\newcommand{\shaking}{\textsf{ShakingHead}}
\newcommand{\nodding}{\textsf{NoddingHead}}
\newcommand{\comehere}{\textsf{ComeHere}}
\newcommand{\maybe}{\textsf{Maybe}}
\newcommand{\surprise}{\textsf{Surprise}}

\newcommand{\shortNameIU}[1]{\textsf{#1}}
\newcommand{\Ne}{\shortNameIU{Ne}}
\newcommand{\Sh}{\shortNameIU{Sh}}
\newcommand{\Nh}{\shortNameIU{Nh}}
\newcommand{\Ch}{\shortNameIU{Ch}}
\newcommand{\Mb}{\shortNameIU{Mb}}
\newcommand{\criteria}[1]{\textsf{#1}}
\newcommand{\SR}{\criteria{SR}}
\newcommand{\G}{\criteria{G}}
\newcommand{\U}{\criteria{U}}
\newcommand{\HE}{\criteria{HE}}
\newcommand{\N}{\criteria{N}}
\newcommand{\CS}{\criteria{S}}
\newcommand{\speechact}[1]{\textsf{#1}}
\newcommand{\representative}{\speechact{Rep}}
\newcommand{\directive}{\speechact{Dir}}
\newcommand{\commisive}{\speechact{Com}}
\newcommand{\expressive}{\speechact{Exp}}
\newcommand{\declaration}{\speechact{Dec}}

\newcommand{\xparagraph}[1]{\textbf{#1}.}

\graphicspath{{./imgs/}}

\newcommand{\concept}[1]{{#1}} 
\newcommand{\iu}{\concept{IU}}
\newcommand{\caudhe}{\concept{CAUDHE}}
\newcommand{\he}{\caudhe}

\newcommand{\m}{\mathbf{m}}

\hypersetup{
  colorlinks   = true, 
  urlcolor     = blue, 
  linkcolor    = blue, 
  citecolor   = blue 
}

\newcommand{\change}[1]{\textcolor{blue}{#1}}
\newcommand{\javichange}[1]{\textcolor{red}{#1}}
\newcommand{\xchange}[1]{\textcolor{purple}{#1}}
\renewcommand{\xchange}[1]{#1}

\renewcommand{\figureautorefname}{Fig.}

%% file: statement.tex
\vfill
\begin{figure*}
{\LARGE
© 2022 IEEE.  Personal use of this material is permitted.  Permission from IEEE must be obtained for all other uses, in any current or future media, including reprinting/republishing this material for advertising or promotional purposes, creating new collective works, for resale or redistribution to servers or lists, or reuse of any copyrighted component of this work in other works.
}
\vfill
\vspace{2cm}
\vfill
\begin{center}
\fbox{\Large
DOI: \href{https://doi.org/10.1109/LRA.2022.3180442}{10.1109/LRA.2022.3180442}
}
\end{center}
\end{figure*}
\vfill

%% file: abstract.tex
Non-verbal communication plays a particularly important role in a wide range of scenarios in Human-Robot Interaction (HRI). Accordingly, this work addresses the problem of human gesture recognition. In particular, we focus on head and eye gestures, and adopt an egocentric (first-person) perspective using eyewear cameras. We argue that this egocentric view \xchange{may offer} a number of conceptual and technical benefits over scene- or robot-centric perspectives.

A motion-based recognition approach is proposed, which operates at two temporal granularities. Locally, frame-to-frame homographies are estimated with a convolutional neural network (CNN). The output of this CNN is input to a long short-term memory (LSTM) to capture longer-term temporal visual relationships, which are relevant to characterize gestures.


Regarding the configuration of the network architecture, one particularly interesting finding is that using the output of an internal layer of the homography CNN increases the recognition rate with respect to using the homography matrix itself.
While this work focuses on action recognition, and no robot or user study has been conducted yet, the system has been designed to meet real-time constraints. 
The encouraging results suggest that the proposed egocentric perspective is viable, and this proof-of-concept work provides novel and useful contributions to the exciting area of HRI.

%% file: introduction.tex
\section{INTRODUCTION}

With the advances in Social Robotics, Human-Robot Interaction (HRI) is an increasingly important area of research. Much work has been conducted on the \emph{robot} side of the interaction so as to generate gaze and body motions which are expressive~\cite{Avdic21dis}, natural\xchange{,} and effective for interaction~\cite{Liu12hri} and collaborations~\cite{Stahl18petra,Terzioglu20hri}, and correspond well to human speech~\cite{Ondras20tc,Ishi18ral}. However, given the importance of non-verbal cues in communication, a successful interaction requires the proper analysis of \emph{human} action~\cite{Ji20tcsvt} as well. While head and eye-based gestures have been investigated in the broader context of human-computer interaction \xchange{(HCI)}~\cite{nuk+16,Drews07mobility,Spakov12ubicomp,Spakov14etra,Hueber20mobileHCI}, less efforts have arguably been devoted in the HRI domain, for instance, to assist and improve human-robot task performance~\cite{Breazeal05iros,Fujii18mia}. 

\xchange{
In HCI, head gestures can be useful for interacting with mobile devices. For instance, touch input can be enriched with slight head motions for shortcut commands~\cite{Hueber20mobileHCI}. Combining gaze and head allows for more natural and accurate interaction~\cite{Spakov12ubicomp}.
Based on smart glasses, gaze and head can be combined following the paradigm of selection-by-looking followed by manipulation-by-moving~\cite{nuk+16}. 
Since small objects can be hard to select using only gaze, subtle head movements can provide additional cues for more accurate hands-free pointing~\cite{Spakov14etra}.
As a further recent evidence, a synergistic use of gaze and head movements can provide users with more flexibility for gaze-based point-and-select tasks~\cite{Sidenmark19uist}. 
Although these concepts and techniques are useful and, to some extent, also applicable in HRI, most research in HCI focuses on interaction with \emph{on-screen} objects. In contrast, gesture in HRI should consider wider and more natural interaction with a \emph{physical} robot and, optionally, with \emph{in-scene} objects. 
Notwithstanding this kind of differences, a common conceptual framework for gestures~\cite{Carfi21toc} in the more general field of human-machine interaction may increase cross-fertilization of otherwise separate research efforts.}
 
In this work, the use of visual-based egocentric perspective for head and eye gesture recognition in the context of HRI is proposed. We argue that the first-person view (FPV) \xchange{may offer} benefits over a third-person view (TPV), as follows: (1) the hard problem of localizing and segmenting human body parts is naturally circumvented; (2) no external (scene or robot) cameras are required; (3) no need that cameras point in \xchange{}specific direction, so the robot might respond to user gestures even if it is not facing the person or is unaware of their presence; (4) the robot does not need to be nearby (it can be in another room in a building or even far away); (5) multiple robots can potentially be interacted with using a single eyewear device. 
\xchange{Therefore, scenarios where the egocentric perspective can be helpful include initiating a dialogue when the robot is not yet ready; interacting with scene objects, where gaze cues are relevant to understand or disambiguate head gestures (e.g. ``grasp \emph{this}" command); in contexts where robot-centric or scene-centric vision are either not possible (e.g. the robot has no camera, or it is remotely operated, or in outdoors interaction), or affordable (e.g. surveillance-like cameras not available in every room).}
There are certainly also some drawbacks such as that the eyewear devices may be uncomfortable to wear, and they might be less cost-effective if multiple persons have to interact with a shared robot. Occlusions may affect both FPV and TPV, although differently, so a redundant system with both types of cameras can perform synergistically and overcome this limitation.

Despite the potential advantages of the egocentric-based head and eye gesture recognition, this problem has been scarcely addressed in HRI. Exceptions include a system to request and guide a robot to find some object~\cite{Zhang13icorr}, helping navigation to wheelchair users with limited hand mobility~\cite{Kutbi21jhir}, or predicting user intention from gaze in grasping tasks~\cite{Kim19sr} and hand-held robots~\cite{Stolzenwald19iros}. Most of these works have in common that they seek to assist disabled people. While these are very important target users to consider, we believe the egocentric perspective can benefit a larger general population, in particular when it comes to interacting with robots in our daily-life environments. 
\xchange{In fact, human gestures recognition can assist human workers in collaborative tasks in industrial settings~\cite{Liu18ijie}, and play an important role for effective and natural human-robot communication in a wide range of applications in social robotics~\cite{Breazeal16SRchapter}, such as education or entertainment. At least in some of those contexts, an egocentric vision perspective represents a valuable alternative or complement to the mainstream robot-centric and scene-centric visual sensors. For instance, when the human subject is holding some objects, smart glasses can capture useful head and eye cues for hands-free interaction~\cite{Park21access}. Human-based egocentric sensors are also helpful in controlling remote robots such as as aerial ones~\cite{Bentz19icra}.}






To address the problem of egocentric head and eye gesture recognition, we propose a motion-based, computer vision system which operates at two temporal levels of the incoming egocentric visual streams. First, frame-to-frame motions are analyzed with a convolutional neural network (CNN) for homography estimation. The output of this network is used as input to a long short-term memory (LSTM) so as to capture longer-range temporal dependencies required to characterize gestures.
LSTMs have recently been proposed in the context of human action recognition, for example using 3D skeleton data encoding parts and joints of the human body~\cite{Li16eccv,son+18,car+19}, or in multimodal approaches~\cite{Islam21ral}.

The contributions of this work are: 
(1)~An egocentric perspective for head and eye gesture recognition for HRI. 
\xchange{This contrasts with the mainstream work which uses robot- or scene-centric vision, and hence further work in this area is promoted.}
(2)~The exploration of an homography estimation network and its joint use with LSTMs. 
\xchange{Although the combination of CNN and LSTMs is not new, the study of a recent homography estimation network, and the fusion of head and eye information are novel aspects.}
(3)~Design decisions and experimental work that provide additional insights for interested practitioners to develop similar systems. \xchange{The good computational and recognition performance are encouraging, and foster further investigation, for which some possibilities are suggested.}


%% file: methodology.tex
\section{METHODOLOGY}

To address the problem of head and eye gesture recognition, an eyewear device with a world camera (W) and one or two eye cameras (E) is {assumed}.

\subsection{Gestures}
\label{s:gestures}

Some studies propose generic gestures for deictic hand gestures~\cite{Cochet19lat}, grasp types~\cite{Feix16thms}, or combining hand gestures and language in the context of HRI~\cite{Matuszek14aaai}, but nothing was found regarding head and eye gestures, specifically for the context of HRI, that could guide our work.

{When gestures-to-commands mappings are arbitrary, like geometric shapes~\cite{Craig16mesa}, the gestures need to be learned and remembered. This issue can be addressed by using on-screen drawings that induce gaze patterns~\cite{Majaranta19etra}, or make user's gaze follow object contour~\cite{Jungwirth18etra}. While valuable, these approaches have some limitations such as that gestures may still be unnatural, or executing them may require some skill, and the need of a display hinders its applicability in some contexts, like many HRI scenarios. More generally speaking, there are a number of factors affecting the usability of gaze~\cite{Istance17chi} and head gestures~\cite{Kutbi21jhir}, which are important to consider in many practical settings.}

{In this work}, the following criteria for head and eye gestures were considered: (1)~oriented to social robots; (2)~general (not application-dependent); (3)~universal (easily understandable by users); (4)~natural (avoiding gestures which are hard to perform or memorize); and (5)~supportive (to express common communication needs).

With these criteria in mind, an initial catalog of interaction units (\iu{s}) was proposed (\autoref{t:catalogue}). The \iu{s} have been classified according to the taxonomy of illocutionary acts~\cite{taxonomy}, which includes \emph{representatives} (\representative) for assertions, \emph{directives} (\directive) for requests or commands, \emph{commissives} for promises, \emph{expressives} (\expressive) for psychological states, and \emph{declarations}. Although this proposal is not definitive and this work only explores a subset of it (the boldfaced \iu{s}, {illustrated in \autoref{f:IUs} and in} videos in the folder `gestures' of Supplementary material),
 we believe it is a good starting point that can guide further work.

{The five head gestures (nodding, turning left/right, and tilting left/right) considered in~\cite{Spakov12ubicomp} resemble our general-purpose gestures. 
With respect to~\cite{Zhang13icorr}, our work shares the idea of using eyewear to communicate with a robot, and exploits the head movements, even though with different purposes and procedures.
The human-robot interaction scenarios where the proposed system might be useful, share commonalities to the kind of non-verbal communication situations considered in~\cite{Brock20roman}, with the difference that that work is based on hand gestures rather than head-eyes. As an example, one specific usage case for the proposed gesture set might be one question-based educational or entertainment game with yes/no answers: after some conversation initiation (\comehere), the robot may ask questions for which the human interactant may respond affirmatively (\nodding), negatively (\shaking), or show some hesitation (\maybe); the robot may also provide some information for which the user may react by displaying understanding (\nodding), disagreement (\shaking), or surprise (\surprise), which might be used by the robot to guide subsequent interaction.} 

\begin{table}[t]
\caption{Proposed \iu\ catalog. The \iu{s} in bold are followed by the identifier used in this paper. The type of \iu\ and involved eyewear images are also given}
\label{t:catalogue}
\centering
\begin{tabular}{|l|c|c|}
\hline
\textbf{Interaction unit} & \textbf{Type} & \textbf{Image(s)} \\ \hline
\textbf{Neutral} (\neutral{})             & -                 & --             \\ 
\textbf{Come here} (\comehere{})          & \directive{}      & W \\ 
Go away                                   & \directive{}      & TBD           \\ 
Stop                                      & \directive{}      & TBD           \\ 
Listen! Look! Hey!                        & \directive{}      & TBD           \\ 
\textbf{Head nod} (\nodding{})            & \representative{} & W, E \\ 
Partial nodding                           & \representative{} & W, E \\ 
\textbf{Head shake} (\shaking{})          & \representative{} & W, E \\ 
\textbf{Positive surprise} (\surprise{})  & \expressive{}     & W, E \\ 
Negative surprise                         & \expressive{}     & TBD           \\ 
\textbf{Maybe} (\maybe{})                 & \representative{} & W \\ \hline
\end{tabular}
\end{table}

\def\myheight{1.7cm}
\def\IUsCaption{{Simplified description of interaction units in terms of rotations ($R_i$)  around, or translations ($T_i$) along some axis $i\in \{X, Y, Z\}$ (a). Side-to-side motions (b) are involved in \shaking\ via $R_Y$, and \maybe\ via $R_Z$. Up-and-down motions (c) are involved in \nodding\ and \comehere\ via $R_X$, with some differences (e.g. repetitions in \comehere\ more time-spaced). \surprise\ may involve a quick gentle backwards motion (d) combining $T_Z$ and $R_X$.}}
\begin{figure}[!t]
\centering
\begin{tabular}{ccccc}
\includegraphics[height=\myheight]{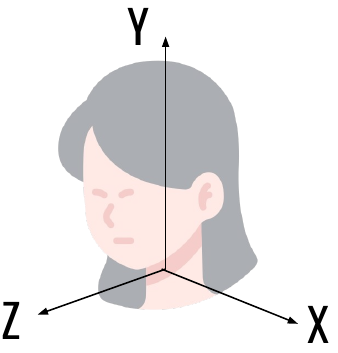} & 
\includegraphics[height=\myheight]{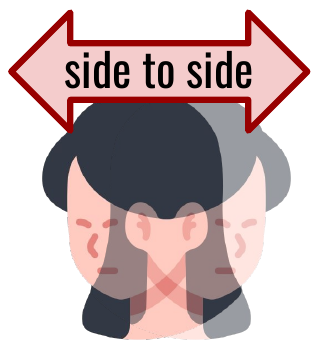}&
\includegraphics[height=\myheight]{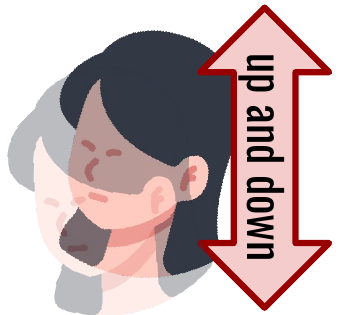} &
\includegraphics[height=\myheight]{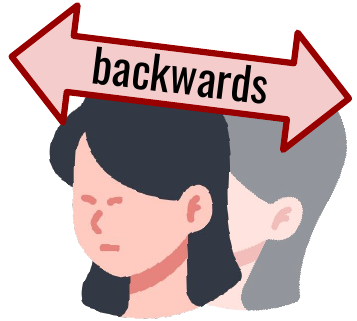} \\
(a) & 
(b) &  
(c) & 
(d) \\ 
\end{tabular}
\caption{\IUsCaption}
\label{f:IUs}
\end{figure}


\subsection{Network architecture and training}

For gesture recognition, we propose a motion-based approach which processes image sequences at two temporal levels (\autoref{fig:network_structure}). First, to analyze global motion between two consecutive video frames, we explore a convolutional network for homography estimation~\cite{caudhe}, which we refer to as \caudhe\ (Content-Aware Unsupervised Deep Homography Estimation). Two key features of this recently proposed network are (1)~its unsupervised nature, which facilitates training since no true homography values are required for each input frame pair, and (2)~its robustness to independently moving objects since the object regions are masked out through an outlier rejector. \caudhe\ addresses some limitations of previous works such as an unsupervised approach~\cite{Nguyen18ral}, whose loss function operates on intensity values rather than on feature space.
Second, to capture long-term visual relationships, the output of \caudhe\ is used as input to a long short-term memory (LSTM). For the size of the output of the LSTM, values $\{16, 32, \ldots, 512\}$ were tested, and 128 was chosen as a good balance between model accuracy and computational cost. A single-layer LSTM was chosen since no benefit was observed with more layers (2, 4 and 8).
Including a batch normalization (BN) layer previous to the input to the LSTM was observed to result in slightly lower \emph{maximum} test accuracy over a set of training runs. However, on average, including this BN layer results in more stable training, quicker convergence, and better \emph{average} accuracy, which we found generally preferable.
After the LSTM, a classification module of a fully connected (FC) layer of $N$ units (the number of \iu{s} considered), is used.

For training, the available pretrained \caudhe's weights\footnote{https://github.com/JirongZhang/DeepHomography} were used, and the weights of the LSTM and the FC layer were supervisedly learned using the true \iu{} labels, with the negative log likelihood as the loss function. 
\autoref{fig:network_structure} illustrates the complete proposed model, which includes two \caudhe\ instances, for the the head and the eye images, respectively. The outputs of these two CNNs are concatenated and input to the LSTM network, as reasoned before.

\def\mywidth{0.35\textwidth}
\begin{figure}[t]
    \centering
    \includegraphics[width=\mywidth]{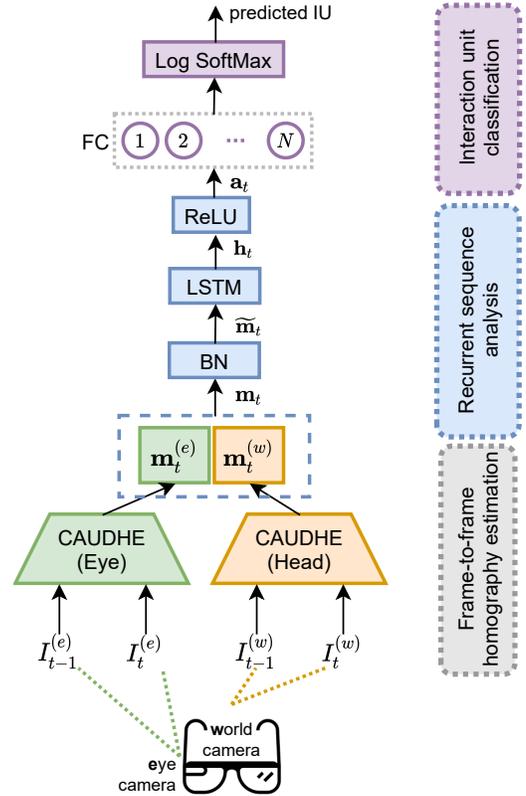}
    \caption{Proposed architecture for head and eye gesture recognition. If only head gestures are considered, only the head \caudhe\ instance is required, and no concatenation of the world and eye output features is needed. }
    \label{fig:network_structure}
    \label{f:model}
\end{figure}

For the actual training of the rest of the model (the sequence analysis and IU classification parts in~\autoref{f:model}), the precomputed features~($\mathbf{m}_t$) from the training videos were split into snippets of length $S=40$ and overlap $O=30$. Batches of $M=32$ snippets were used for training speedup, better performance of the BN layer, and faster convergence.

\subsection{Practical considerations}
\label{s:practical_issues}

Two important design factors for the system to perform efficiently and effectively are the frame rate and the image sizes, since they play important roles in different aspects, 
 and a careful design and proper tradeoff is called for. 
Since \caudhe\ assumes a small baseline ($B$) in the input frame pairs, the higher the frame rate ($f$), the more likely this constraint will be satisfied.
On the other hand, the world camera of the eyewear device can operate at several spatial resolutions (from 1080p Full HD to QVGA), but the highest frame rates are only selectable with lower resolutions of these world images.
Furthermore, higher quality images have a notable GPU memory footprint, often exceeding the available resources. Therefore, since frames will be later resized, it makes little sense to acquire frames at high spatial resolution.

When using low frame rates, we observed blurring artifacts in the world images when fast head movements were performed. These effects were alleviated with $f>30$ fps. Thus, if images are high quality, homography estimation should improve.
However, computational resources impose constraints on both the image size (to prevent memory overflow) and the frame rate (to keep under the limited computational power). Therefore, both spatial and temporal subsampling were required (\autoref{t:camera_config}), and their effects are discussed below (\autoref{s:subsampling}).



\begin{table}[t]
\caption{Final configuration of acquisition (A) and working (W) resolutions after subsampling}
\label{t:camera_config}
\begin{center}
\begin{tabular}{|l|cc|cc|}
\hline
\multirow{2}{*}{\textbf{Camera}} & \multicolumn{2}{|c|}{Image size, $w\times h$ (px)} & \multicolumn{2}{|c|}{frame rate (fps)} \\
\cline{2-5}
 & A & W & A & W \\
\hline
World & $640 \times 480$ & $192\times 144$ ($30\%$) & \multirow{2}{*}{60} & \multirow{2}{*}{Up to 30}\\
Eye   & $192 \times 192$ & $192\times 192$ ($100\%$) &  &  \\
\hline
\end{tabular}
\end{center}
\end{table}








\subsection{Inference and online recognition}
\label{s:online}
For efficiency at inference time, the input video frames are fed into \caudhe\ in batches of $K$ frame pairs for processing $K$ homographies in parallel on the GPU. The $K$ resulting features ($\mathbf{m}_t$) are reshaped and fed to the LSTM as a $K$-length sequence. 
Although predictions of \iu{s} are performed frame-wise, for more robust classification, majority voting {can be} applied to the resulting $K$ predicted IU labels. Here, $K=10$ was chosen as a good tradeoff between computational efficiency and classification delay.

%% file: experiments.tex
\section{EXPERIMENTS}

\subsection{Configuration}

\begin{figure}
\centering
    \includegraphics[width=0.65\columnwidth]{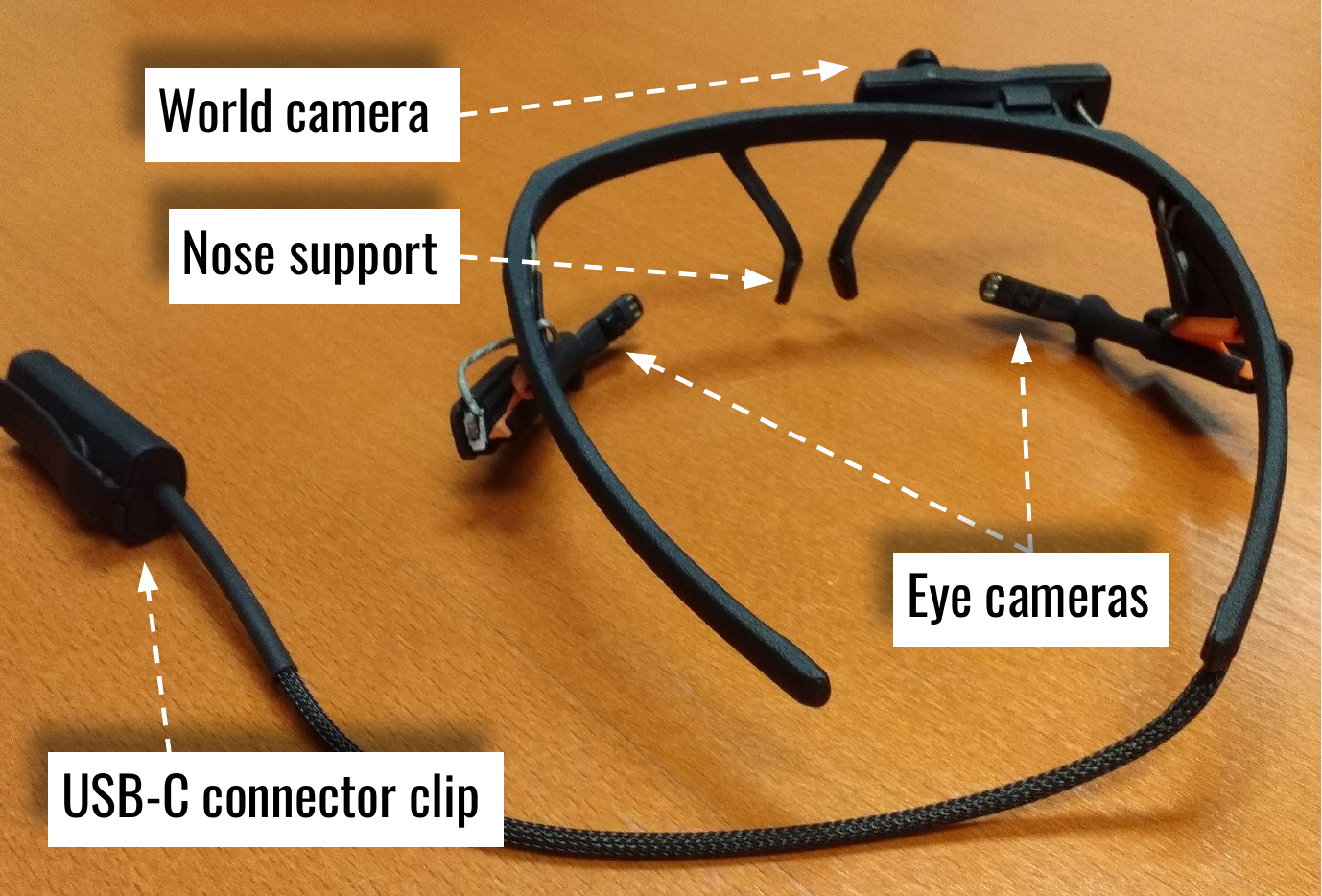}
    \caption{{Pupil Core headset. Please refer to~\url{pupil-labs.com/products/core} and~\cite{pupil} for technical details}}
\label{f:pupil}
\end{figure}
For the purposes of this work, the images from the world camera and from the right-eye camera of the Pupil Core headset~\cite{pupil}{~(\autoref{f:pupil})} were used.

To the best of our knowledge, no public egocentric gesture dataset exists that meets the required criteria (\autoref{s:gestures}). Therefore, we collected a dataset with 10 minutes of video, 
half indoors (close-to-camera walls and objects) and half outdoors (distant background), with about 150 seconds (10 videos $\times$ 15 seconds/video) from each of four participants\footnote{This work has been approved by the deontological board of Universitat Jaume I (file code ``CD/70/2021'', date June 17th, 2021). Human subjects participated after their informed consent.}. 
{The participants, volunteer master students, were each briefly instructed about the gestures, without much detail, so that performance was as naturally as possible. Some IUs required additional guidance (see~\autoref{s:discussion}). Prior to the actual recording, a calibration step (about 10--20 seconds long) per participant plus a few trial gestures were performed.}
Each of the 40 videos (10 videos per participant) contains 3--4 non-overlapping repetitions of a single non-neutral \iu. The \iu\ statistics {of the dataset} are summarized in \autoref{t:iu_stats}. Video frames were {manually} annotated with the corresponding true \iu\ for supervised learning. Since \surprise\ turned out to be a particularly subtle \iu, it is {considered in \autoref{f:surprise_eyes} in} \autoref{s:world_eye} {to separately study the effect of including the eye images}, while the rest of the reported results consider the other five~\iu{s}. {All tests include the world and eye images except in~\autoref{s:deep_features}, which focuses on the effect of deep features.} {Conditions are summarized in~\autoref{t:summ}.}
\def\H{H}
\def\D{D}
\def\A{A}
\def\Am{A${}^-$}
\def\B{B}
\def\W{W}
\def\E{E}
\def\EW{\W+\E}
\begin{table}
\caption{{Summary of experimental conditions.}
{Image (W=World, E=Eye), Features (H=Homography, D=Deep), IUs (\A=all, \Am=all but \surprise, \B=two IUs)}}
\label{t:summ}
\centering
\def\HD{\ref{f:homo_vs_deep}}
\def\WE{\ref{f:word_eye}}
\def\SU{\ref{f:surprise_eyes}}
\def\FR{\ref{f:frame_rates}}
\def\BI{\ref{f:binary}}
\def\GE{\ref{f:generalisation}}
\def\CM{\ref{f:final_confusion_matrix}}
\def\TD{\ref{f:final_time_diagrams}}
\begin{tabular}{|l|c|c|c|c|c|}
\hline
Aspect $\blacktriangledown$ \textbackslash\; Figure $\blacktriangleright$ & \HD    & \WE         & \SU         & \BI & \FR, \GE, \CM, \TD \\
\hline
Images    & \W    & \multicolumn{3}{|c|}{\{\W, \E, \EW\}} & \multicolumn{1}{|c|}{W+E} \\  \hline
Features  & \{\H, \D\}  & \multicolumn{4}{|c|}{\D} \\
 \hline
IUs       & \multicolumn{2}{|c|}{\Am} & \A & \B & \multicolumn{1}{|c|}{\Am} \\
\hline

\end{tabular}
\end{table}

{The \neutral\ IU is more represented in the dataset, which results in a significant class imbalance situation. To deal with it, the \neutral\ class as under-sampled (a ratio of training snippets from this class are discarded) so as to reduce the imbalance level. This was proven very effective and used in all reported experiments.}

To account for the stochastic nature of training neural models, 30 repetitions were performed, over which performance statistics on the validation sets are reported. 
Experiments were run on modest low-cost hardware (PC, 8 GB, 1.8 GHz, GeForce MX250 with 2 GB GPU). Most software was developed using the Python ecosystem (NumPy, pandas, scikit-learn, PyTorch, etc.).

\begin{table}[t]
\caption{Distribution of instances of \iu{s} in the dataset}
\label{t:iu_stats}
\centering
\begin{tabular}{|l|c|c|c|}
\cline{1-4}
\multicolumn{1}{|c|}{\textbf{IU}} & \textbf{Length (s)} & \textbf{\%} & \textbf{\# instances} \\ \hline
\multicolumn{1}{|l|}{\neutral{}}  & 423.77  & 70.63 & N/A      \\ 
\multicolumn{1}{|l|}{\shaking{}}  & 37.38   & 6.23  & 25       \\ 
\multicolumn{1}{|l|}{\nodding{}}  & 34.73   & 5.79  & 26       \\ 
\multicolumn{1}{|l|}{\comehere{}} & 27.13   & 4.52  & 30       \\ 
\multicolumn{1}{|l|}{\maybe{}}    & 46.74   & 7.79  & 30       \\ 
\multicolumn{1}{|l|}{\surprise{}} & 30.25   & 5.04  & 32       \\ \hline
\multicolumn{1}{|l|}{{All IUs}} & {600.00}   & {100} & --       \\ \hline
\end{tabular}
\end{table}

The model was trained for 30 epochs, with Adam optimizer, a learning rate of $5 \cdot 10^{-4}$, weight decay penalty of $10^{-2}$, $\beta_1=0.9$, $\beta_2=0.999$, and PyTorch's \textsl{ReduceLROnPlateau} scheduler. A stratified, \iu\ label-based split of 75\% of training instances was used, except in~\autoref{s:generalization}.

\subsection{Homography vs. deep features}
\label{s:deep_features}

For gesture recognition, an obvious first choice is to use the~8 values of the estimated homography as input to the LSTM ($\mathbf{m}_t$ in \autoref{f:model}). However, we are not concerned on motion estimation itself, but on a qualitative representation of motion that can properly characterize the different interaction units. Therefore, we hypothesized that using the output of a hidden layer of \caudhe\ could be beneficial. In particular, the output of the last 2D average pooling layer of \caudhe\ (512 features) was used as deep features, and compared with the 8-dim homography values. As found by the distribution of the maximum accuracy (\autoref{f:homo_vs_deep}), using the deep features for $\mathbf{m}_t$ turns out to be favorable, which indicates that this qualitative representation of motion is richer and more discriminative than the homography estimates. Similar results were obtained for other performance metrics (precision, recall, and~$F_1$~score).

\def\mywidth{0.5\columnwidth}
\begin{figure}[t]
    \centering
    \includegraphics[width=\mywidth]{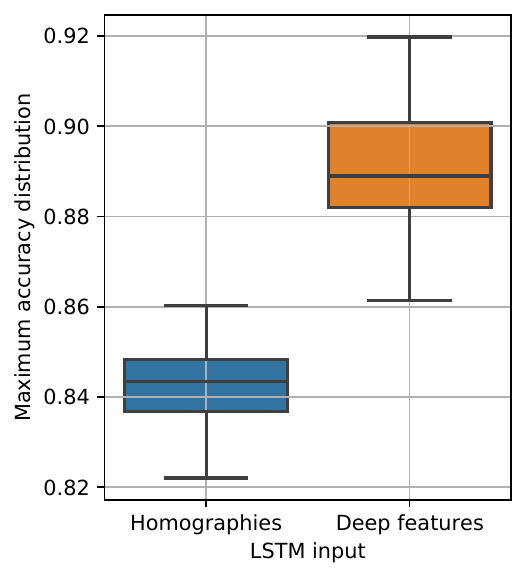}
    \caption{Maximum accuracy with different motion representations used as input to the LSTM}
    \label{f:homo_vs_deep}
\end{figure}

\subsection{Using the world and/or eye images}
\label{s:world_eye}

We evaluated how the world ($I^{(w)}_t$) and eye ($I^{(e)}_t$) images contribute to the recognition performance. It can be observed (\autoref{f:word_eye}) that the performance is {lower} when only eye images are considered, which is not surprising given that the interaction units are heavily head-based. 
\def\mywidth{0.70\columnwidth}
\begin{figure}[t]
    \centering
    \includegraphics[width=\mywidth]{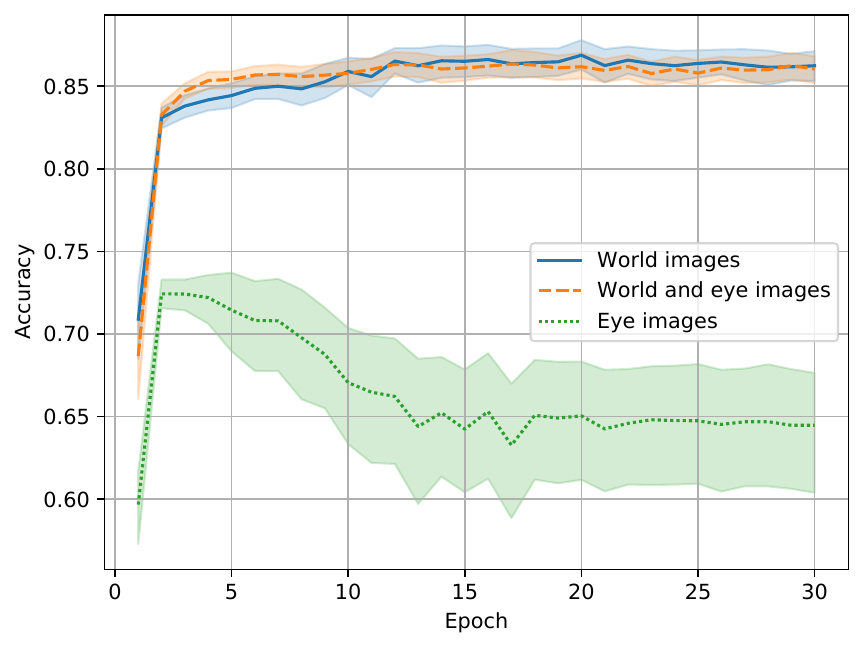}
    \caption{Average accuracy with world and/or eye images} 
    \label{f:word_eye}
\end{figure}
Although the joint consideration of the world and eye images does not improve performance, it is interesting that the recognition does not degrade even when eye images do not bring clearly distinctive information. This is a promising indication that the model might leverage on eye images for capturing commonalities and individual differences for a larger set of images from more participants and repetitions, or when more interaction units, some relying more on eye cues (blinks, winks, subtle motions, etc.), are considered. 
As an illustrative example, the recognition of the \surprise\ \iu{} benefits significantly from including the eye images, with an $F_1$ score about twice that of using the world images only (\autoref{f:surprise_eyes}).

\def\mywidth{0.70\columnwidth}
\begin{figure}[t]
    \centering
    \includegraphics[width=\mywidth]{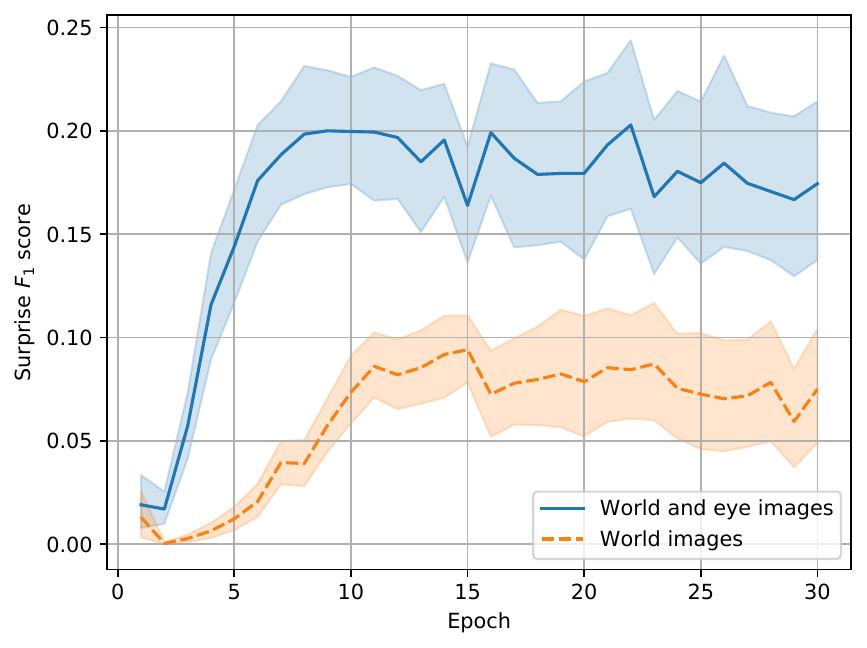}
    \caption{Average $F_1$ for \surprise\ with and without eye images}
    \label{f:surprise_eyes}
\end{figure}


{As a controlled experiment, we considered a weighted concatenation $[\alpha_e \m_t^{(e)}, \alpha_w \m_t^{(w)}]$, and trained the network separately for \nodding\ vs \neutral\ and \surprise\ vs \neutral\, with $\alpha_e=1$ and varying $\alpha_w\in[0,2]$, 
so as to include $\alpha_e\ll \alpha_w$, $\alpha_e \simeq \alpha_w$ and $\alpha_e\gg \alpha_w$. Results~(\autoref{f:weighted_concat}) show that the convergence is faster and the final performance better with bigger $\alpha_w$ for \nodding, whereas $\alpha_w$ has smaller relevance for the performance for \surprise. This is a sign that eye images provide higher discriminative cues than world images for some IUs such as \surprise.}

\def\mywidth{0.85\columnwidth}
\begin{figure}[t]
    \centering
    \includegraphics[width=\mywidth]{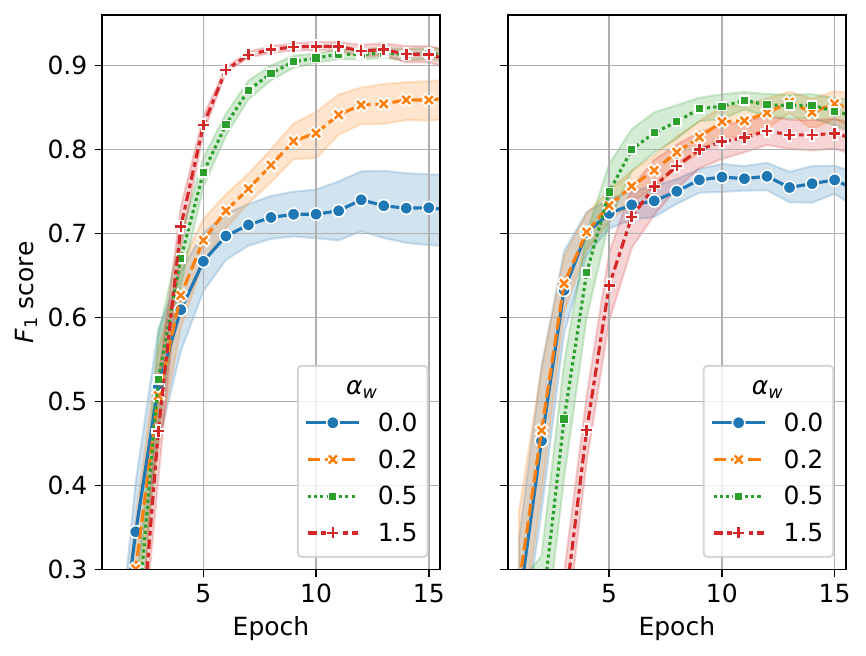}
    \caption{{Performance for varying concatenation weight $\alpha_w$. Left: \nodding\ vs \neutral. Right: \surprise\ vs \neutral}}
    \label{f:weighted_concat}
    \label{f:binary}
\end{figure}
 
\subsection{Effect of subsampling}
\label{s:subsampling}
Regarding spatial subsampling, it was observed that $192\times 144$ (30\% of the original VGA resolution) worked well in most scenarios. For temporal subsampling, the analysis of the network convergence with different frame rates (\autoref{f:frame_rates}) 
reveals that, although all get to approximately the same recognition accuracy, convergence is faster with higher frame rates. For instance, for $f=10$ fps, more than 10 epochs are required to get an accuracy which is obtained with 5 or less epochs for the other frame rates. The likely reason for this behavior is that the baseline between two consecutive frames is smaller with higher frame rates, and this benefits the homography estimation network. Furthermore, the robustness against smaller frame rates can be attributed to the use of deep features. For the final version of the proposed system, $f=20$ fps is chosen as a good tradeoff solution. 

\def\mywidth{0.7\columnwidth}
\begin{figure}[t]
    \centering
    \includegraphics[width=\mywidth]{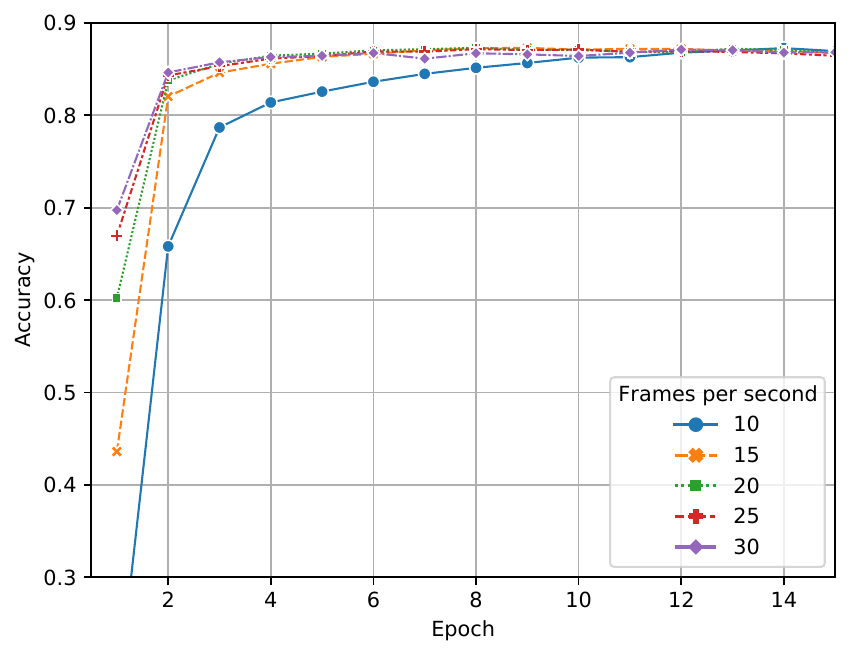}
    \caption{Average accuracy with different frame rates}
    \label{f:frame_rates}
\end{figure}


\subsection{Generalization ability}
\label{s:generalization}
\begin{figure}[t]
    \centering
    \includegraphics[width=0.80\columnwidth]{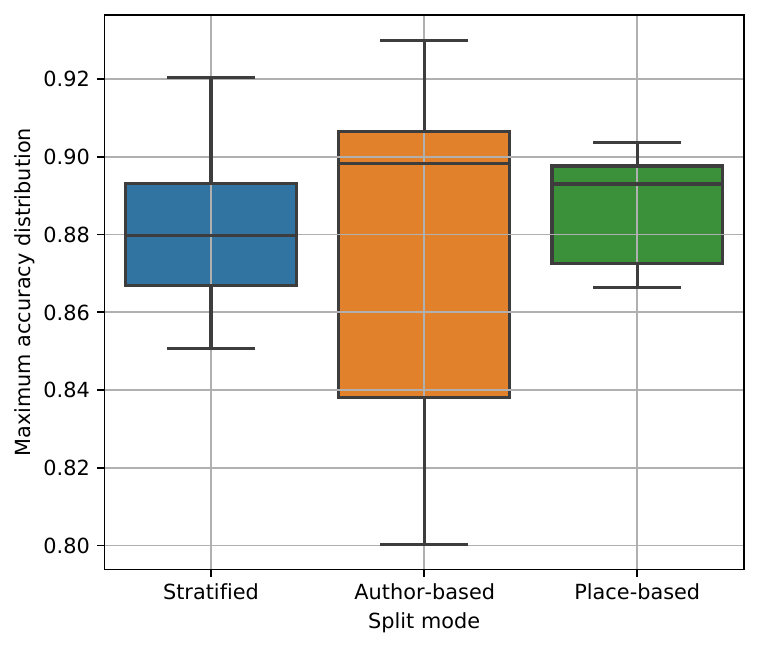}
    \caption{Distribution of the maximum accuracy on validation videos during training, for different dataset split methods}
    \label{f:generalisation}
\end{figure}

To evaluate the ability of the proposed model to recognize \iu{s} performed by different subjects (authors) or against different backgrounds (places), we compare the stratified dataset split to the author-based splits (i.e. leaving-one-author out), and place-based splits (i.e. indoor and outdoor recordings are separately considered in the training and test splits). Results (\autoref{f:generalisation}) indicate that in the author-based case the maximum accuracy has larger variability than in the stratified case, since different subjects perform the gestures differently, but in most training repetitions the maximum accuracy was nonetheless high (about 84\%). Regarding the place-based split, the performance metric has lower variability and similar or better average value than the stratified case, which is a good sign of the ability of \caudhe\ to deal with scenes with backgrounds at different depths, and possibly the positive effect of using deep features. 
Furthermore, being motion-based, the recognition approach is not biased by the different backgrounds, which might be distracting with appearance-based approaches.
Taken these results together, and given the limited data available, it can be argued that the model generalizes reasonably well to different scenes and users.

\subsection{Overall performance}
To understand which \iu{s} are generally better recognized, or with which other \iu{s} they can be misrecognized, it is useful to analyze the confusion matrix (\autoref{f:final_confusion_matrix}). It is revealed that \neutral{} is recognized well by our model. \maybe\ is recognized with a high accuracy, and is the best recognized ``non-neutral'' \iu. \shaking{} is recognized fine as well, but is confused sometimes with \neutral{}. Finally, \nodding{} and \comehere{} can also be mistaken, as they both imply vertical head movements, and their difference is somehow subtle and hard to distinguish. As the \comehere{} movements are more subtle, it is more frequently confused with \neutral{}.
{Globally, this performance can be understood by the similarity among groups of the \iu{s} (\autoref{f:IUs}), and by other factors such as the variability in the performance by different participants, and the limited dataset size (\autoref{s:discussion}).}
{As a representative performance summary, an overall accuracy of 90\% resulted from averaging the peak validation accuracy over 40 training runs, for these five IUs, using deep features and both the world and eye images.}
\input confMatFig4-31

\subsection{Online prediction}
Finally, in addition to global performance metrics, it is insightful to see the predicted \iu{s} over time. 
{To that end, the network is fed with consecutive pairs of frames, and its output (the predicted IU) at each time step is reported as is, unfiltered.}
Results (\autoref{f:final_time_diagrams}) indicate that \shaking\ (\autoref{f:final_time_diagram_5}) and \maybe\ (\autoref{f:final_time_diagram_8}) are generally the best recognized interaction units, and the model tends to label incorrectly only the frames close to transitions.
\nodding\ (\autoref{f:final_time_diagram_3}) is confused with \comehere\ and even \maybe, but the recognition performance is still acceptable. 
In the case of \comehere\ (\autoref{f:final_time_diagram_7}), although the overall accuracy is low (actions are often misrecognized as \nodding), the correct IU is detected on the onset of the IU, which is the most important part in practical situations.
A video demonstration is given in folder `online' in the Supplementary material, 
{where majority voting (\autoref{s:online}) is applied for robustness.}


\def\mywidth{\linewidth}
\def\subfigwidth{0.9\columnwidth}
\begin{figure}[!t]
    \centering
\begin{tabular}{c}
    \begin{subfigure}[b]{\subfigwidth}
        \centering
        \includegraphics[width=\mywidth]{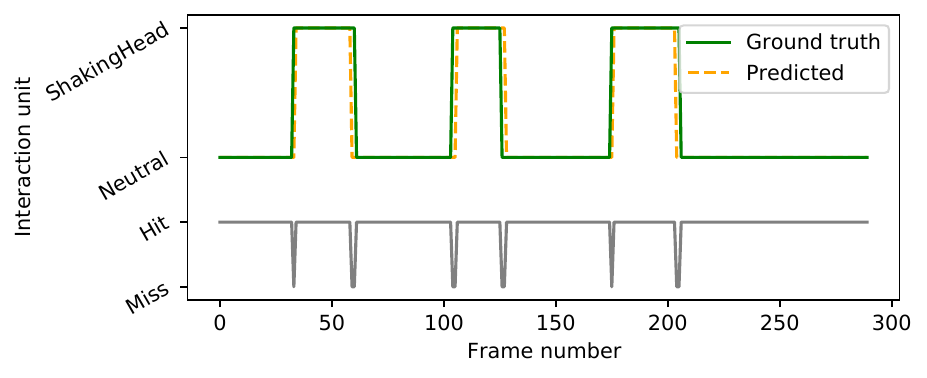}
        \caption{\shaking}
        \label{f:final_time_diagram_5}
    \end{subfigure} \\
    \begin{subfigure}[b]{\subfigwidth}
        \centering
        \includegraphics[width=\mywidth]{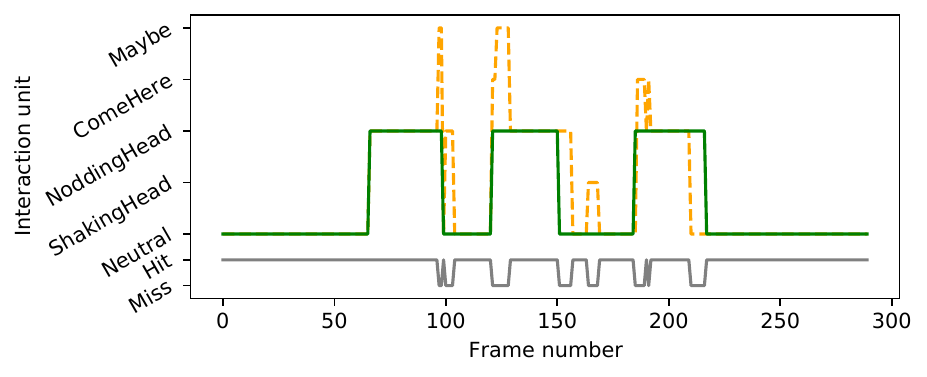}
        \caption{\nodding}
        \label{f:final_time_diagram_3}
    \end{subfigure} \\ 
    \begin{subfigure}[b]{\subfigwidth}
        \centering
        \includegraphics[width=\mywidth]{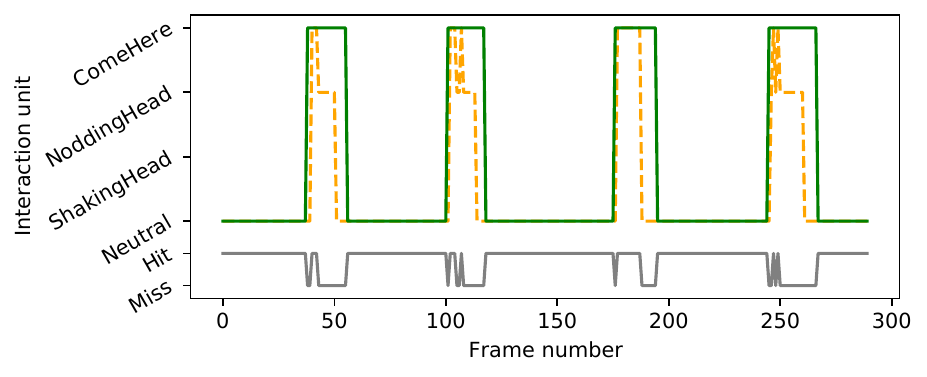}
        \caption{\comehere}
        \label{f:final_time_diagram_7}
    \end{subfigure} \\ 
    \begin{subfigure}[b]{\subfigwidth}
        \centering
        \includegraphics[width=\mywidth]{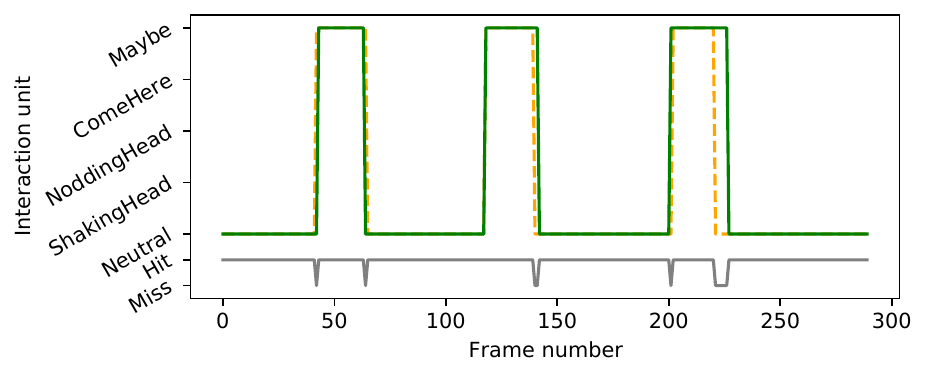}
        \caption{\maybe}
        \label{f:final_time_diagram_8}
    \end{subfigure}
    \end{tabular}
    \caption{Time diagrams of some validation videos involving various \iu{s}. The bottom gray line helps to appreciate correct recognitions (``hit") and misrecognitions (``miss")}
    \label{f:final_time_diagrams}
\end{figure}

%% file: confMatFig4-31.tex
\begin{figure}[t]
\centering
\def\myConfMat{{
{2122,   31,   12,   34,   14},  
{  11,  158,    0,    0,    0},  
{  46,    0,  136,   30,    2},  
{   5,    0,    9,   72,    0},  
{  18,    0,    8,    0,  192},  
}}

\def\classNames{{"Ne","Sh","Nh","Ch","Mb"}} 

\def\numClasses{5} 

\def\myScale{1.0} 
\begin{tikzpicture}[
    scale = \myScale,
    ]

\tikzset{vertical label/.style={rotate=90,anchor=east}}   
\tikzset{diagonal label/.style={rotate=45,anchor=north east}}

\foreach \y in {1,...,\numClasses} 
{
    \node [anchor=east] at (0.4,-\y) {\pgfmathparse{\classNames[\y-1]}\pgfmathresult}; 
   
    \foreach \x in {1,...,\numClasses}  
    {
    \def\totSamples{0}
    \foreach \ll in {1,...,\numClasses}
    {
        \pgfmathparse{\myConfMat[\ll-1][\x-1]}   
        \xdef\totSamples{\totSamples+\pgfmathresult} 
    }
    \pgfmathparse{\totSamples} \xdef\totSamples{\pgfmathresult}  
    
    \begin{scope}[shift={(\x,-\y)}]
        \def\mVal{\myConfMat[\y-1][\x-1]} 
        \pgfmathtruncatemacro{\r}{\mVal}   %
        \pgfmathtruncatemacro{\p}{round(\r/\totSamples*100)}
        \coordinate (C) at (0,0);
        \ifthenelse{\p<50}{\def\txtcol{black}}{\def\txtcol{white}} 
        \node[
            draw,                 
            text=\txtcol,         
            align=center,         
            fill=black!\p,        
            minimum size=\myScale*10mm,    
            inner sep=0,          
            ] (C) {\r\\\textbf{\p}\%};     
        \ifthenelse{\y=\numClasses}{
        \node [] at ($(C)-(0,-4.75)$) 
        {\pgfmathparse{\classNames[\x-1]}\pgfmathresult};}{}
    \end{scope}
    }
}
\coordinate (yaxis) at (-0.5,0.5-\numClasses/2);  
\coordinate (xaxis) at (0.5+\numClasses/2, \numClasses-4.8); 
\node [vertical label] at (yaxis) {Predicted IU};
\node []               at (xaxis) {True IU};
\end{tikzpicture}

\caption{Confusion matrix. \Ne=\neutral, \Sh=\shaking, \Nh=\nodding, \Ch=\comehere, \Mb=\maybe}
\label{f:final_confusion_matrix}
\end{figure}

%% file: discussion.tex
\section{DISCUSSION}
\label{s:discussion}
\xparagraph{\xchange{Data aspects}} The proposed approach for egocentric motion-based gesture recognition has been proven reliable in the collected dataset. Further work would be required to test the system with more interaction units and more data. 
\xchange{Class imbalance was addressed by under-sampling the majority class. It could be explored whether additional mechanisms~\cite{Johnson19jbd} bring some further benefit.}

\xparagraph{\xchange{Gestures}} 
At collection time, some \iu{s} (\nodding\ and \shaking) simply required some prompts from the experimenter to ensure natural performance, but some other \iu{s} required some hints (to avoid ``shrugging shoulders" for \maybe, since that would not involve head or eyes), or suggestions for some exaggeration (\surprise\ was too subtle otherwise). In the future, elicitation studies might help waive such indications for more natural performance. \xchange{The gesture catalog was proposed towards meeting criteria such as universality and usability. However, some gestures are subject to differences in context of use and cultural variations, with implications in their meaning~\cite{Matsumoto13jnvb}, social acceptability~\cite{Alallah18vrst,Hossain21mum}, etc. 
Therefore, adaptation of some gestures to different applications or world regions may be required.}

\xparagraph{\xchange{Network design}}
Fusion techniques other than feature concatenation for combining world and eye images could be explored.
Possibly, an alternative, local encoding of eye images (e.g. optic-flow based) would be preferable over the current global motion assumption. 
An open interesting question is whether performance can be improved if the homography estimation network is trained or fine-tuned with \xchange{gesture-specific} sequences instead of using the available pretrained weights.

\xparagraph{\xchange{Egocentric view}} 
\xchange{Since the eyewear device is light-weight, it can be reasonably comfortable to wear, mostly in comparison to virtual-reality devices. It is a also preferable over head-mounted cameras because it allows the recognition of gestures where eye cues are important, including deictic gestures. 
In addition, having a view closer to the wearer's eyes is generally desirable.
Beyond gestures, gaze provides proxies for cognitive states (e.g. attention) which can be relevant for the interaction.}
Although the human-egocentric perspective is promising and potentially useful in some practical scenarios, this view can be complemented\xchange{, where possible,} with robot-egocentric and scene-centric visual sensors (e.g. for face, upper-body or full-body analysis), thus aiming at more robust recognition and wider applicability.
\xchange{While this work focuses on human gesture recognition, the interesting related problem of human-to-robot motion retargeting~\cite{Zhang22ral,Khalil21roman} might be explored from an alternative, egocentric perspective, for generating robot (head) movements.}

%% file: conclusions.tex
\section{CONCLUSIONS}

A motion-based head and eye gesture recognition framework has been proposed that leverages on egocentric visual data from an eyewear device. The approach relies on the combination of an homography estimation convolutional neural network for frame-to-frame motion characterization, and a long short-term memory for capturing longer-range visual dependencies. 
Five general-purpose interaction units have been shown to be recognized with \xchange{reasonable high accuracy} while the system can run in real time with commodity hardware. Moreover, the system performs equally well in a range of frame rates (e.g. 20--30 fps). These are very encouraging results for practical human-robot interaction requirements. One of the most interesting findings is that using the output of a hidden layer of the homography network increases the recognition performance over using the actual output of raw homography values.